\def\year{2021}\relax
\documentclass[letterpaper]{article} 
\usepackage[switch]{lineno}  %
\usepackage{aaai21}  
\usepackage{times}  
\usepackage{helvet} 
\usepackage{courier}  
\usepackage[hyphens]{url}  
\usepackage{graphicx} 
\usepackage{natbib}  
\usepackage{caption} 
\usepackage[linesnumbered,ruled,vlined]{algorithm2e}
\usepackage{xcolor}
\usepackage{amsfonts}
\usepackage[fleqn]{amsmath}
\usepackage{amstext}
\usepackage{subcaption}

\title{Enhanced Regularizers for Attributional Robustness}

\author {
    Anindya Sarkar,\textsuperscript{\rm 1}
    Anirban Sarkar, \textsuperscript{\rm 1}
    Vineeth N Balasubramanian \textsuperscript{\rm 1} \\
}
\affiliations {
    \textsuperscript{\rm 1} Indian Institute of Technology, Hyderabad \\
    anindya.sarkar@cse.iith.ac.in, cs16resch11006@iith.ac.in, vineethnb@ith.ac.in
}

\begin{document}
\maketitle

\vspace{-5pt}
\begin{abstract}
Deep neural networks are the default choice of learning models for computer vision tasks. Extensive work has been carried out in recent years on explaining deep models for vision tasks such as classification. However, recent work has shown that it is possible for these models to produce substantially different attribution maps even when two very similar images are given to the network, raising serious questions about trustworthiness. To address this issue, we propose a robust attribution training strategy to improve attributional robustness of deep neural networks. Our method carefully analyzes the requirements for attributional robustness and introduces two new regularizers that preserve a model's attribution map during attacks. Our method surpasses state-of-the-art attributional robustness methods by a margin of approximately 3\% to 9\% in terms of attribution robustness measures on several datasets including MNIST, FMNIST, Flower and GTSRB.
\end{abstract}

\vspace{-5pt}
\section{Introduction}
\label{Introduction}
\label{sec:Introduction}
In order to deploy machine learning models in safety-critical applications like healthcare and autonomous driving, it is desirable that such models should reliably explain their predictions. As neural network architectures become increasingly complex, explanations of the model's prediction or behavior are even more important for developing trust and transparency to end users. For example, if a model predicts a given pathology image to be benign, then a doctor might be interested to investigate further to know what features or pixels in the image led the model to this classification.

\begin{figure*}[t]
\centering
  \includegraphics[width=0.9\textwidth,height=4.5cm]{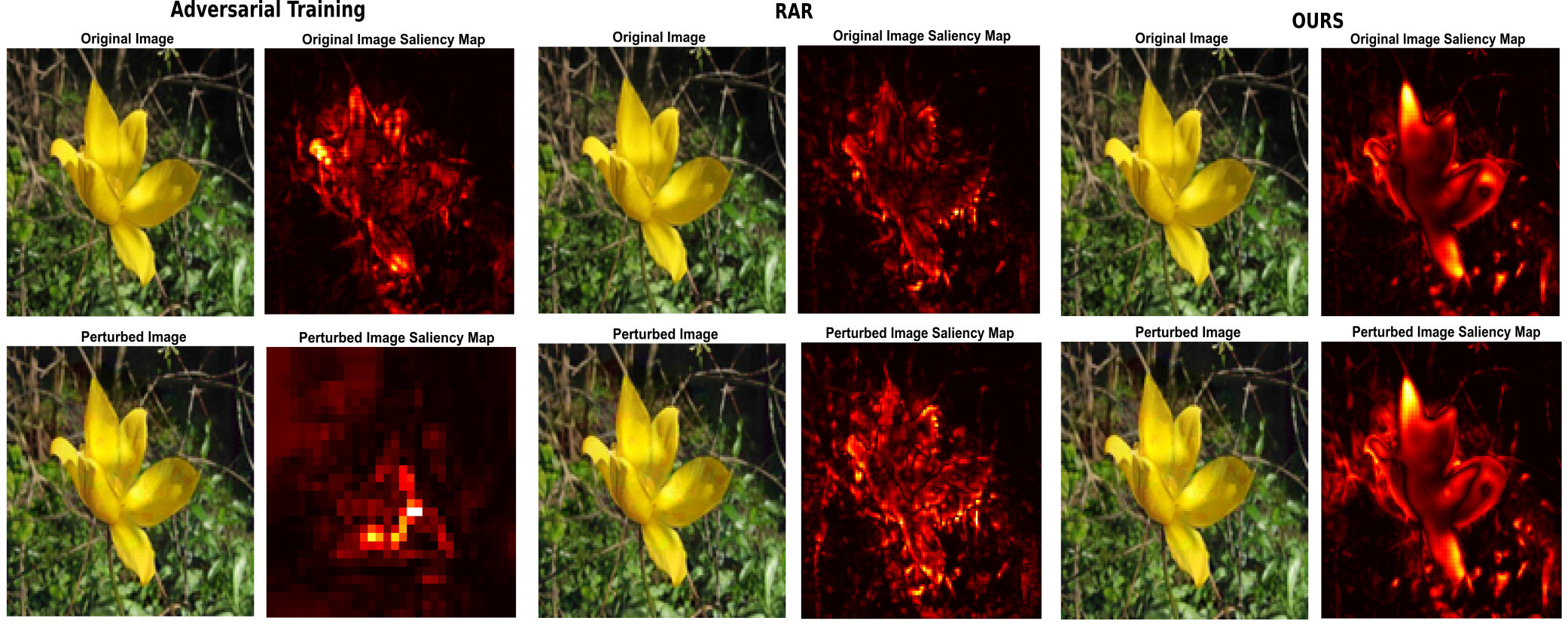}
  \vspace{-3pt}
  \caption{Original image and human-imperceptible attributional attacks (same predicted class) and their corresponding saliency maps under: \textit{(Left)} adversarial training \cite{madry2017towards}; \textit{(Middle)} RAR \cite{chen2019robust}; and \textit{(Right)} our proposed robust attribution training for a test sample from Flower dataset. Note that the attacks/perturbed images in each column are different, since the attack is carried out based on the model trained on that particular method.}
  \vspace{-7pt}
  \label{fig:Introduction Image}
\end{figure*}

Though there exist numerous efforts in the literature on constructing adversarially robust models \cite{qin2019adversarial,wang2020improving,zhang2019theoretically,madry2017towards,chan2019jacobian,sarkar2020enforcing,xie2019feature}, surprisingly very little work has been done in addressing issues in robustness of the explanations generated by a model. 
One aspect of genuineness of a model can be in producing very similar interpretations for two very similar human-indistinguishable images where model predictions are the same. \cite{ghorbani2019interpretation} demonstrated the possibilities to craft changes in an image which are imperceptible to a human, but can induce huge change in attribution maps without affecting the model’s prediction. Hence building robust models, against such attacks proposed in \cite{ghorbani2019interpretation}, is very important to increase faithfulness of such models to end users. Fig \ref{fig:Introduction Image} visually explains the vulnerability of adversarial robust models against attribution-based attacks and how attributional training (\`{a} la adversarial training) addresses this to a certain extent. Such findings imply the need to explore effective strategies to improve attributional robustness.

The limited efforts until now for attributional training rely on minimizing change in attribution due to human imperceptible change in input \cite{chen2019robust} or maximizing similarity between input and attribution map \cite{kumari2019harnessing}. We instead propose a new methodology for attributional robustness that is based on empirical observations. Our studies revealed that attributional robustness gets negatively affected when: (i) 
an input pixel has a high attribution for a negative class (non-ground truth class) during training; (ii) an attribution map corresponding to the positive or true class 
is uniformly distributed across the given image, instead of being localized on a few pixels; or  and 
(iii) change of attribution, due to an human imperceptible change in input image (without changing predicted class label), is higher for a pixel with low attribution than for a pixel with high attribution (since this leads to significant changes in attribution). 
Based on these observations, we propose a new training procedure for attributional robustness that addresses each of these concerns and outperforms existing attributional training methods. Our methodology is  inspired by the rigidity (and non-amorphous nature) of objects in images and instigates the fact that number of true class pixels are often small compared to total number of pixels in an image, resulting in a non-uniform (or skewed) pixel distribution of the true class attribution across spatial locations in the image. Complementarily, for the most confusing negative class, ensuring the attribution or saliency map is \textit{not localized} helps indirectly improve the localization of the attribution of the true class. 

The key contributions of our work can be summarized as follows: (i) We propose a \textit{class attribution-based contrastive regularizer} for attributional training which forces the true class attribution to assume a skewed shape distribution, replicating the fact that few input pixels attribute highly compared to other pixels for a true class prediction. We also drive pixels of the attribution map corresponding to the negative class to behave uniformly (equivalent to not localizing) through this regularizer;
(ii) We also introduce an \textit{attribution change-based regularizer} to weight change in attribution of a pixel due to an indistinguishable change in the image (both the aforementioned contributions have not been attempted before to the best of our knowledge);
(iii) We provide detailed experimental results of our method on different benchmark datasets, including MNIST, Fashion-MNIST, Flower and GTSRB, and obtain state-of-the-art results for attributional robustness across these datasets.

An illustrative preview of the effectiveness of our method is shown in Fig \ref{fig:Introduction Image}. The last column shows the utility of our method, where the attribution/saliency map (we use these terms interchangeably in this work) seems stronger, and minimally affected by the perturbation. The middle column shows a recent state-of-the-art \cite{chen2019robust}, whose attribution shows signs of breakdown from the perturbation.


\section{Related Work}
\label{Related Work}
\label{sec:Related Work}
We divide our discussion of related work into subsections that capture earlier efforts that are related to ours from different perspectives.

\noindent \textbf{Adversarial Robustness:} The possibility of fooling neural networks by crafting visually imperceptible images was first shown by \cite{szegedy2013intriguing}. Since then, we have seen extensive efforts over the last few years in the same direction. \cite{goodfellow2014explaining} introduced one-step Fast Gradient Sign Method (FGSM) attack which was followed by more effective iterative attacks such as \cite{kurakin2016adversarial}, PGD attack \cite{madry2017towards}, Carlini Wagner attack \cite{carlini2017towards}, Momentum Iterative attack \cite{dong2017discovering}, Diverse Input Iterative attack \cite{xie2019improving}, Jacobian-based saliency map approach \cite{papernot2016limitations}, etc. A parallel line of work has also emerged on finding strategies to defend against stronger adversarial attacks such as Adversarial Training \cite{madry2017towards}, Adversarial Logit Pairing \cite{kannan2018adversarial}, Ensemble Adversarial Training \cite{tramer2017ensemble}, Parsevals Network \cite{cisse2017parseval}, Feature Denoising Training \cite{xie2019feature}, Latent Adversarial Training \cite{kumari2019harnessing}, Jacobian Adversarial Regularizer \cite{chan2019jacobian}, Smoothed Inference \cite{nemcovsky2019smoothed}, etc. The recent work of \cite{zhang2019theoretically} explored the trade-off between adversarial robustness and accuracy.

\noindent \textbf{Interpretability Methods:} The space of work on robust attributions is based on generating neural network attributions which is itself an active area of research. These methods have an objective to compute the importance of input features based on the prediction function's output. Recent efforts in this direction include gradient-based methods \cite{attr2013gradient,attr2016inputgradient,sundararajan2017axiomatic}, propagation-based techniques \cite{bach2015pixel,shrikumar2017learning,attr2018ebackprop,nam2019relative} or perturbation-based methods \cite{attr2016perturb1,attr2018perturb2,attr2016perturb3lime}. Another recent work \cite{dombrowski2019explanations} developed a smoothed explanation method that can resist manipulations, while our work aims to develop a training method (not explanation method) that is resistant to attributional attacks. Our work is based on integrated gradients \cite{sundararajan2017axiomatic} which has often been used as a benchmark method \cite{chen2019robust,singh2019benefits} and is theoretically well-founded on axioms of attribution, and shown empirically strong performance.

\noindent \textbf{Attributional Attacks and Robustness:} The perspective that neural network interpretations can be broken by modifying the saliency map significantly with imperceptible input perturbations - while preserving the classifier’s prediction - was first investigated recently in \cite{ghorbani2019interpretation}. While the area of adversarial robustness is well explored, little progress has been made on attributional robustness i.e. finding models with robust explanations.
\cite{chen2019robust} recently proposed a training methodology which achieves current state-of-the-art attributional robustness results.
It showed that attributional robustness of a model can be improved by minimizing the change in input attribution w.r.t an imperceptible change in input. Our work is closest to this work, where an equal attribution change on two different input pixels are treated equally, irrespective of the original pixel attribution. This is not ideal as a pixel with high initial attribution may need to be preserved more carefully than a pixel with low attribution. \cite{chen2019robust} has another drawback as it only considers input pixel attribution w.r.t. true class, but doesn't inspect the effect w.r.t. other negative classes. We find this to be important, and have addressed both these issues in this work. Another recent method \cite{singh2019benefits} tried to achieve better attributional robustness by encouraging the observation that a meaningful saliency map of an image should be perceptually similar to the image itself. This method fails specifically for images where the true class contains objects darker than the rest of the image, or there are bright pixels anywhere in the image outside the object of interest. In both these cases, enforcing similarity of saliency to the original image objective shifts the attribution away from the true class in this method. We compare our method against both these methods in our experiments.


\vspace{-3pt}
\section{Background and Preliminaries}
\label{sec_bgd_prelims}
Our proposed training method requires computation of input pixel attribution through the Integrated Gradient (IG) method \cite{sundararajan2017axiomatic}, which has been used for consistency and fair comparison to the other closely related methods \cite{chen2019robust, singh2019benefits}. It functions as a technique to provide axiomatic attribution to different input features proportional to their influence on the output. Computation of IG is mathematically approximated by constructing a sequence of images interpolating from a baseline to the actual image and then averaging the gradients of neural network output across these images, as shown below: 
 \begin{equation}
 \label{eqn:ig}
IG_{i}^f(\textbf{x}_{0},\textbf{x}) = (\textbf{x}^{i} - \textbf{x}^{i}_{0})\times \sum_{k=1}^{m} \frac{\partial f(\textbf{x}^{i}_{0} + \frac{k}{m} \times (\textbf{x}^{i} - \textbf{x}^{i}_{0} ))}{\partial \textbf{x}^{i}} \times \frac{1}{m}
\end{equation}
Here $f:\mathbb{R}^n \rightarrow \mathcal{C}$ represents a deep network with $\mathcal{C}$ as the set of class labels, $\textbf{x}_0$ is a baseline image with all black pixels (zero intensity value) and $i$ is the pixel location on input image $\textbf{x}$ for which IG is being computed.
 
\noindent \textbf{Adversarial Attack:}  We evaluate the robustness of our model against two kinds of attacks, viz. Adversarial Attack and Attributional Attack, each of which is introduced herein. The goal of an adversarial attack is to find out minimum perturbation $\delta$ in the input space of $\textbf{x}$ (i.e. input pixels for an image) that results in maximal change in classifier($f$) output. In this work, to test adversarial robustness of a model, we use one of the strongest adversarial attacks, Projected Gradient Descent (PGD) \cite{madry2017towards}, which is considered a benchmark for adv accuracy in other recent attributional robustness methods \cite{chen2019robust,singh2019benefits}. PGD is an iterative variant of Fast Gradient Sign Method (FGSM) \cite{goodfellow2014explaining}. PGD adversarial examples are constructed by iteratively applying FGSM and projecting the perturbed output to a valid constrained space $S$. PGD attack is formulated as follows:
\begin{equation}
    \textbf{x}^{i+1}\> =\> Proj_{\textbf{x}+S}\>(\textbf{x}^{i}\>+\> \alpha {(\nabla_{\textbf{x}} \mathcal{L}(\theta,\textbf{x}^{i},\textbf{y})))}
    \label{eq:PGD}
\end{equation}
Here, $\theta$ denotes the classifier parameters; input and output are represented as \textbf{x} and \textbf{y} respectively; and the classification loss function as $\mathcal{L}(\theta,\textbf{x},\textbf{y})$. Usually, the magnitude of adversarial perturbation is constrained in a $L_p$-norm ball ($p \in \{ 0,2,\infty \}$) to ensure that the adversarially perturbed example is perceptually similar to the original sample. Note that $\textbf{x}^{i+1}$ denotes the perturbed sample at $(i+1)^{th}$ iteration.

\noindent \textbf{Attributional Attack:} The goal of an attributional attack is to devise visually imperceptible perturbations that change the interpretability of the test input maximally while preserving the predicted label. To test attributional robustness of a model, we use Iterative Feature Importance Attack (IFIA) in this work. As \cite{ghorbani2019interpretation} convincingly demonstrated, IFIA helps generate minimal perturbations that substantially change model interpretations, while keeping their predictions intact. The IFIA method is formally defined as below: 
\begin{equation}
\vspace{-2pt}
    \operatorname{\arg\max}_\delta\> D(I(\textbf{x};f), I(\textbf{x}+\delta;f))
\label{eq:attributional attack}
\end{equation}
\begin{center}
    \text{ subject to:} $|| \delta ||_{\infty} \leq \epsilon$\\
    \text{ such that:} $\max\>f(\textbf{x};\theta) = \max\>f(\textbf{x}+\delta;\theta)$
\end{center}

Here, $I(\textbf{x},f)$ is a vector of attribution scores over all input pixels when an input image $\textbf{x}$ is presented to a classifier network $f$ parameterized by $\theta$.
$D(I(\textbf{x};f),I(\textbf{x}+\delta;f))$ measures the dissimilarity between attribution vectors $I(\textbf{x};f)$ and $I(\textbf{x}+\delta;f)$. 
In our work, we choose $D$ as Kendall's correlation computed on top-$k$ pixels as in \cite{ghorbani2019interpretation}. We describe this further in the Appendix due to space constraints.



    

    

\vspace{-3pt}
\section{Proposed Methodology}
\label{sec:Methodology}
We now discuss our proposed robust attribution training strategy in detail, which: (i) enforces restricting true class attribution as a sparse heatmap and the negative class attribution to be a uniform distribution across the entire image; (ii) enforces the pixel attribution change caused by an imperceptible perturbation of the input image to consider the actual importance of the pixel in the original image. Both these objectives are achieved through the use of a regularizer in our training objective. The standard multiclass classification setup is considered where input-label pairs $(\textbf{x},\textbf{y})$ are sampled from training data distribution $\mathcal{D}$ with a neural network classifier $f$, parametrized by $\theta$. Our goal is to learn $\theta$ that provides better attributional robustness to the network.

\noindent \textbf{Considering Negative Classes:} We observe that ``good'' attributions generated for a true class form a localized (and sparse, considering the number of pixels in the full image) heatmap around a given object in the image (assuming an image classification problem setting). On the other hand, this implies that we'd like the most confusing/uncertain class attribution to not be localized, viz. i.e. resemble a uniform distribution across pixels in an image. As stated earlier, this hypothesis is inspired by the rigidity (and non-amorphous nature) of objects in images. 
To this end, we define the \textit{Maximum Entropy Attributional Distribution} as a discrete uniform distribution in input pixel space as $P_{ME}$, where attribution  score of each input pixel is equal to $\frac{1}{number \>of\>pixels}$. We also define a \textit{True Class Attributional Distribution ($P_{TCA}$)} as a distribution of attributions over input pixels for the true class output, denoted by $f^{TCI}$, when provided the perturbed image as input. Note that attributions are implemented using the IG method (as described in Sec \ref{sec_bgd_prelims}), and $P_{TCA}$ hence averages the gradients of the classifier's true class output when input is varied from $\textbf{x}_{0}$ to $\textbf{x}^{\prime}$. We also note that IG is simply a better estimate of the gradient, and hence can be computed w.r.t. every output class (we compute it for the true class here). Here $\textbf{x}_0$ is a baseline reference image with all zero pixels, and $\textbf{x}^{\prime}$ represents the perturbed image. $\textbf{x}^{\prime}$ is chosen randomly within an  $l_{\infty}$-norm $\epsilon$-ball around a given input $\textbf{x}$. We represent $P_{TCA}$ then as:
\vspace{-3pt}
\begin{equation}
P_{TCA}(\textbf{x}) = \text{softmax}\big(IG^{f_{TCI}}_{\textbf{x}}(\textbf{x}_{0}, \textbf{x}^{\prime})\big)
\label{eq_P_TCA_computation}
\vspace{-2pt}
\end{equation}
where $IG^{f_{TCI}}(\cdot,\cdot)$ is computed for every pixel in $\textbf{x}$, and the softmax is applied over all $P$ pixels in $\textbf{x}$, i.e. $\text{softmax}(\textbf{u}_{i}) = \frac{\exp(\textbf{u}_i)}{\sum_{j \in P} \exp(\textbf{u}_j)}$.

In a similar fashion, we define a \textit{Negative Class Attributional Distribution ($P_{NCA}$)}, where IG is computed for the most confusing negative class (i.e. class label with second highest probability) in a multi-class setting, or simply the negative class in a binary setting. $P_{NCA}$ is given by:
\begin{equation}
P_{NCA} = \text{softmax}(IG^{f_{NCI}}_{\textbf{x}}(\textbf{x}_{0}, \textbf{x}^{\prime}))
\label{eq_P_NCA_computation}
\end{equation}

We now define our \textit{Class Attribution-based Contrastive Regularizer (CACR)} as: 
\begin{equation}
\mathcal{L}_{CACR} = KL(P_{ME}|| P_{NCA}) - KL(P_{ME}|| P_{TCA})
\label{eq_CACR_computation}
\end{equation}
where $KL$ stands for KL-divergence. We show how CACR is integrated into the overall loss function to minimize, later in this section. CACR enforces a skewness in the attribution map, corresponding to the true class, across an input image through the "$-KL(P_{ME}|| P_{TCA})$" term, and a uniform attribution map corresponding to the most confusing negative class through the "$KL(P_{ME}|| P_{NCA})$" term. The skewness in case of the true class forces the learner to focus on a few pixels in the image. This regularizer induces a contrastive learning on the training process, which is favorable to attributional robustness, as we show in our results.


\vspace{-10pt}
\paragraph{Enforcing Attribution Bounds:} If a pixel has a positive (or negative) attribution towards true class prediction, it may be acceptable if a perturbation makes the attribution more positive (or more negative, respectively). In other words, we would like the original pixel attribution to serve as a lower bound for a positively attributed pixel, or an upper bound for a negatively attributed pixel. If this is violated, it is likely that the attribution map may change. To implement this thought, 
we define $\mathcal{A}(\textbf{x})$ as a \textit{base attribution} i.e. computed using standard IG method attribution w.r.t the true class for the input image $\textbf{x}$, given by:
\begin{equation}
\mathcal{A}(\textbf{x}) = IG^{f_{TCI}}_{\textbf{x}}(\textbf{x}_{0}, \textbf{x})
\end{equation}
Similarly, we define $\nabla\mathcal{A}(\textbf{x})$ as the change in attribution w.r.t the true class given the perturbed image, i.e. (a similar definition is also used in \cite{chen2019robust}):
\begin{equation}
\nabla\mathcal{A}(\textbf{x}) = IG^{f_{TCI}}_{\textbf{x}}(\textbf{x}, \textbf{x}^{\prime}) = IG^{f_{TCI}}_{\textbf{x}}(\textbf{x}_{0}, \textbf{x}^{\prime}) - IG^{f_{TCI}}_{\textbf{x}}(\textbf{x}_{0}, \textbf{x})
\end{equation}
The abovementioned desideratum necessitates that the sign of every element of $\mathcal{A} \odot \nabla\mathcal{A}$, where $\odot$ is the elementwise/Hadamard product, be maintained positive across all pixels. To understand better, let us consider the $i^{\text{th}}$ pixel in $\mathcal{A}$ to be positive (negative). This implies that the $i^{\text{th}}$ pixel is positively (negatively) affecting classifier’s true class prediction. In such a case, we would like the $i^{\text{th}}$ component of $\nabla\mathcal{A}$ also to be positive (negative), i.e. it further increases the magnitude of attribution in the same direction (positive/negative, respectively) as before.

However, even when the $\text{sign}(\mathcal{A} \odot \nabla\mathcal{A})$ is positive for a pixel, we argue that an equal amount of change in attribution on a pixel with higher base attribution is more costly compared to a pixel with lower base attribution, i.e. we also would want the magnitude of each element in $\mathcal{A} \odot \nabla\mathcal{A}$ to be low, in addition to the overall sign being positive. 

Our second regularizer, which we call \textit{Weighted Attribution Change Regularizer (WACR)}, seeks to implement the above ideas. This is achieved by considering two subsets of pixels in a given input image $\textbf{x}$: a set of pixels $P_1$ for which $\text{sign}(\mathcal{A} \odot \nabla\mathcal{A})$ is negative, i.e. sign($\mathcal{A}$) is not the same as sign($\nabla\mathcal{A}$); and a set of pixels $P_2$ for which sign($\mathcal{A}$) is same as sign($\nabla\mathcal{A}$). We then minimize the quantity below:
\begin{equation}
\mathcal{L}_{WACR} = S(\nabla\mathcal{A})\bigg|_{p \in P_1} +
     S(\mathcal{A} \odot \nabla\mathcal{A})\bigg|_{p \in P_2}
\label{eq_WACR_computation}
\end{equation}
where we choose $S(\cdot)$ can be any size function, which we use as $L_1$-norm in this work, and $p \in P_1 \cup P_2$ is a pixel from the image $\textbf{x}$. In Eqn \ref{eq_WACR_computation}, we note that:
\vspace{-3pt}
\begin{itemize}
\setlength\itemsep{-0.1em}
    \item the first term attempts to reduce attribution change in pixels where sign($\mathcal{A}$) is not the same as sign($\nabla\mathcal{A}$). We argue that this reduction in $\nabla\mathcal{A}$ is not required for pixels in $P_2$, since an attribution change helps reinforce correct attributions for pixels in $P_2$.
    \item the second term attempts to lower the change in attribution more in pixels with higher base attribution. We argue that this is not required for pixels in $P_1$, since bringing down the attribution change irrespective of the base attribution is the focus for $P_1$.
\end{itemize}



\begin{table*}[h]
\vspace{-2mm}
\footnotesize
    \begin{subtable}{0.99\linewidth}
        \centering
        \begin{minipage}{.5\linewidth}
		\begin{tabular}{|p{1.50cm}|p{1.10cm}|p{1.20cm}|p{0.9cm}|p{0.95cm}|}
		\hline \hline
		Method & Nat. acc. & Adv. acc. & Top-K & Kendall \\
		\hline \hline
        Natural & \textbf{86.76}\% & 0.00\% & 8.12\% & 0.4978 \\
        \hline
        Madry et al. & 83.82\% & 41.91\% & 55.87\% & 0.7784 \\
        \hline
        RAR & 82.35\% & 47.06\% & 66.33\% & 0.7974 \\
        \hline
        \textbf{Ours} & 83.09\% & \textbf{51.47}\% & \textbf{69.50\%} & \textbf{0.8121} \\
        \hline \hline
		\end{tabular}
		\caption{Results on Flower dataset}
		\label{tab:Flower main}
        \end{minipage}%
        \begin{minipage}{.5\linewidth}
		\begin{tabular}{|p{1.50cm}|p{1.10cm}|p{1.20cm}|p{0.9cm}|p{0.95cm}|}
		\hline  \hline
		Method & Nat. acc.& Adv. acc.&
		Top-K & Kendall \\
		\hline \hline
        Natural & \textbf{90.86}\% & 0.01\% & 39.01\% & 0.4610 \\
        \hline
        Madry et al. & 85.73\% & \textbf{73.01}\% & 46.12\% & 0.6251 \\
        \hline
        RAR & 85.44\% & 70.26\% & 72.08\% & 0.6747 \\
        \hline
        \textbf{Ours} & 85.45\% & 71.61\% & \textbf{81.50\%} & \textbf{0.7216} \\
        \hline \hline
		\end{tabular}
		\caption{Results on Fashion-MNIST dataset}
		\label{tab:Fashion-MNIST main}
        \end{minipage}
    \end{subtable}
    \hfill
    \begin{subtable}{0.99\linewidth}
        \centering
        \begin{minipage}{.5\linewidth}
		\begin{tabular}{|p{1.50cm}|p{1.10cm}|p{1.20cm}|p{0.9cm}|p{0.95cm}|}
		\hline \hline
		Method & Nat. acc.& Adv. acc.&
		Top-K & Kendall \\
		\hline \hline
        Natural & \textbf{99.17}\% & 0.00\% & 46.61\% & 0.1758 \\
        \hline
        Madry et al. & 98.40\% & \textbf{92.47}\% & 62.56\% & 0.2422 \\
        \hline
        RAR & 98.34\% & 88.17\% & 72.45\% & 0.3111 \\
        \hline
        \textbf{Ours} & 98.41\% & 89.53\% & \textbf{81.00\%} & \textbf{0.3494} \\
        \hline \hline
		\end{tabular}
		\caption{Results on MNIST dataset}
		\label{tab:MNIST main}
        \end{minipage}%
        \begin{minipage}{.5\linewidth}
		\begin{tabular}{|p{1.50cm}|p{1.10cm}|p{1.20cm}|p{0.9cm}|p{0.95cm}|}
		\hline \hline
        Method & Nat. acc.& Adv. acc.&
		Top-K & Kendall \\
		\hline \hline
        Natural & \textbf{98.57}\% & 21.05\% & 54.16\% & 0.6790 \\
        \hline
        Madry et al. & 97.59\% & \textbf{83.24}\% & 68.85\% & 0.7520 \\
        \hline
        RAR & 95.68\% & 77.12\% & 74.04\% & 0.7684 \\
        \hline
        \textbf{Ours} & 97.57\% & 82.33\% & \textbf{77.15\%} & \textbf{0.7889} \\
        \hline \hline
		\end{tabular}
		\caption{Results on GTSRB dataset}
		\label{tab:GTSRB main}
        \end{minipage}
    \end{subtable}
\end{table*}

\begin{figure*}[t]
\centering
  \includegraphics[width= 0.79\textwidth,height=3.2cm]{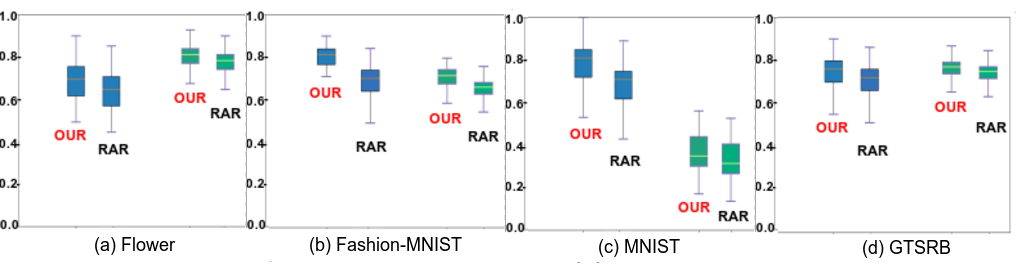}
  \vspace{-7pt}
  \caption{\textit{(Best viewed in color)} Variance in attribution robustness metrics, Top-$k$ intersection score and Kendall's correlation, for proposed method and RAR over all test samples on different datasets. Blue=Top-$k$ intersection score; Green=Kendall's correlation. Our method achieves significant improvement on both metrics over RAR for all datasets, while the variance is fairly similar to variance in RAR}
  \label{fig:All Box Results}
\end{figure*}

\begin{table*}[!htb]
\footnotesize
    \begin{minipage}{.5\linewidth}
        \begin{center}
		\begin{tabular}{|p{1.50cm}|p{1.10cm}|p{1.20cm}|p{0.9cm}|p{0.95cm}|}
		\hline \hline
		    Method & Nat. acc. & Adv. acc. & Top-K & Kendall \\
			\hline \hline
			Natural & \textbf{93.91}\% & 0.00\% & 38.22\% & 0.5643 \\
            \hline
            Madry et al. & 92.64\% & 69.85\% & 80.84\% & 0.8414 \\
            \hline
            Singh et al. & 93.21\% & 33.08\% & 79.84\% & 0.8487 \\
            \hline
            Ours & 90.31\% & \textbf{74.26}\% & \textbf{95.50\%} & \textbf{0.9770}\\
            \hline \hline
		\end{tabular}
	\end{center}
	\vspace{-8pt}
		\caption{Results on Flower dataset (WRN 28-10 architecture)}
		\label{tab:Flower WRN}
    \end{minipage}%
    \begin{minipage}{.5\linewidth}
        \begin{center}
		\begin{tabular}{|p{1.50cm}|p{1.10cm}|p{1.20cm}|p{0.9cm}|p{0.95cm}|}
		\hline \hline
			Method & Nat. acc. & Adv. acc. & Top-K & Kendall \\
			\hline \hline
            Natural & \textbf{99.43}\% & 19.9\% & 68.74\% & 0.7648 \\
            \hline
            Madry et al. & 98.36\% & \textbf{87.49}\% & 86.13\% & 0.8842 \\
            \hline
            Singh et al. & 98.47\% & 84.66\% & 91.96\% & 0.8934 \\
            \hline
            Ours & 98.41\% & 85.17\% & \textbf{92.14\%} & \textbf{0.9502} \\
            \hline \hline
		\end{tabular}
	\end{center}
	\vspace{-8pt}
		\caption{Results on GTSRB dataset (WRN 28-10 architecture)}
		\label{tab:GTSRB WRN}
    \end{minipage}
    \vspace{-6mm}
\end{table*}

\noindent \textbf{Overall Optimization:}
We follow an adversarial training approach \cite{madry2017towards} to train the model. Adversarial training is a two-step process: an (i) Outer minimization; and an (ii) Inner maximization. The inner maximization is typically used to identify a suitable perturbation that achieves the objective of an attribution attack, and the outer minimization seeks to use the regularizers described above to counter the attack. We describe each of them below:

\noindent \textit{Outer Minimization:} Our overall objective function for the outer minimization step is given by:
\vspace{-3pt}
\begin{equation}
\min_{\theta} l_{CE}(\textbf{x}^{\prime},\textbf{y};\theta) + \lambda \big(\mathcal{L}_{CACR} + \mathcal{L}_{WACR} \big)
\label{Eq:Outer minimization}
\vspace{-3pt}
\end{equation}
where $l_{CE}$ is the standard cross-entropy loss used for the multi-class classification setting. We use $\lambda$ as a common weighting coefficient for both regularizers, and use $\lambda = 1$ for all the experiments reported in this paper. We show effects of considering different $\lambda$ values on our proposed method in in Sec \ref{sec:Ablation}.
As $P_{ME}$, $P_{NCA}$ and $P_{TCA}$ are all discrete distributions, we calculate $\mathcal{L}_{CACR}$ as:
\begin{equation}
    \sum_{i=1}^{P} P_{ME}(\textbf{x}_{i}) \odot \Big( \log \frac{P_{ME}(\textbf{x}_{i})}{P_{NCA}(\textbf{x}_{i})} -  \log \frac{P_{ME}(\textbf{x}_{i})}{P_{TCA}(\textbf{x}_{i})}\Big)
\label{eq_CACR_calculation}
\end{equation}
where $P$ corresponds to the total number of pixels in the input image, as before. $P_{NCA}(\textbf{x}_{i})$ corresponds to the 1st order partial derivative of neural network output (corresponding to most confusing negative class) w.r.t the $i^{\text{th}}$ input pixel. Similarly, $P_{TCA}(\textbf{x}_{i})$ corresponds to the 1st order partial derivative of neural network output (corresponding to true class) w.r.t the $i^{\text{th}}$ input pixel.

\noindent \textit{Inner Maximization:} In order to obtain the attributional attack, we use the following objective function:
\begin{equation}
 \max_{\textbf{x}^{\prime} \in N(\textbf{x},\epsilon)}  l_{CE}(\textbf{x}^{\prime},\textbf{y};\theta) + S(\nabla\tilde{\mathcal{A}})
 \label{Eq:Inner maximization}
\end{equation}
\begin{center}
    where $\nabla\tilde{\mathcal{A}} = IG^{\mathcal{L}_{TCI}}_{\textbf{x}}(\textbf{x}, \textbf{x}^{\prime})$
\end{center}
Earlier computations of IG were computed w.r.t $f_{TCI}$ or $f_{NCI}$, which were the softmax outputs of the true class and the most confusing negative class respectively. Here, we denote $\tilde{\mathcal{A}}$ to denote the computation of IG using the loss value corresponding to the true class. This is because our objective here is to maximize loss, while our objective was to maximize the true class softmax output in the outer minimization. We use $\mathcal{L}_{TCI}$ as the cross-entropy loss for the true class, and $L_1$-norm as $S(\cdot)$. Since the inner maximization is iterative by itself (and solved before the outer minimization), we randomly initialize each pixel of $\textbf{x}^\prime$ within an $l_{\infty}$-norm ball of $\textbf{x}$ and then iteratively maximize the objective function in Eqn \ref{Eq:Inner maximization}. We avoid the use of $\mathcal{L}_{CACR}$ in our inner maximization, since $\mathcal{L}_{CACR}$ is expensive due to an extra IG calculation w.r.t. negative class, which can increase the cost due to the many iterations in the inner maximization loop.

We note that the proposed method is not an attribution method, but a training methodology that uses IG. When a model is trained using our method, all axioms of attribution will hold for IG by default, as for any other trained model. We also show that our loss function can be used as a surrogate loss of the robust prediction objective proposed by \cite{madry2017towards}. Please refer to Appendix for the proof. An algorithm for our overall methodology is also presented in the Appendix due to space constraints.

\label{sec:Ablation}
\begin{figure*}[t]
\centering
  \includegraphics[width= 0.8\textwidth,height=4.8cm]{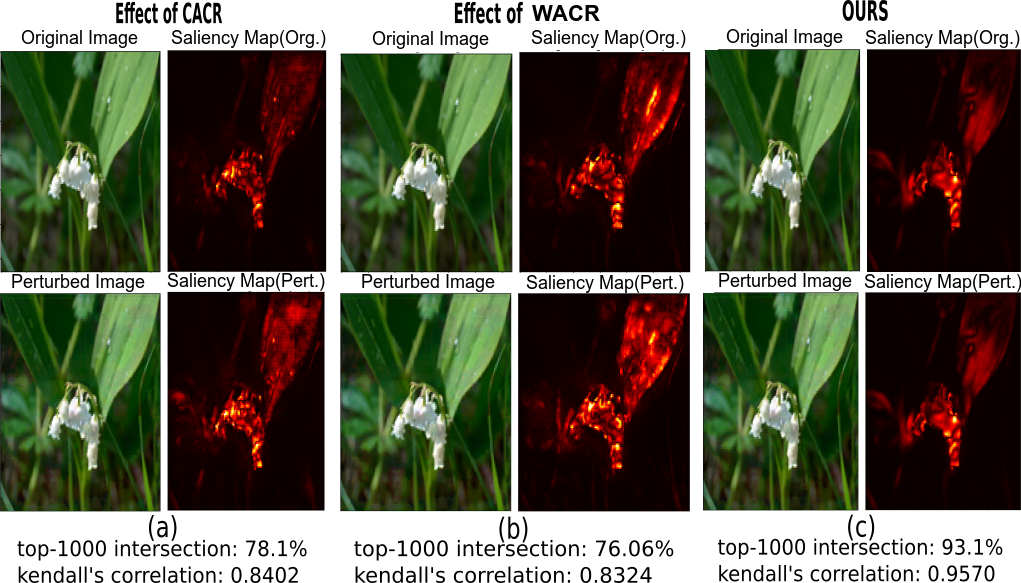}
  \vspace{-4pt}
  \caption{Visualizing effects of different regularization terms. Columns (a), (b), (c) represent saliency maps of original and perturbed image considering only $\mathcal{L}_{CACR}$, only $\mathcal{L}_{WACR}$ and both regularizers respectively in outer minimization (Eqn \ref{Eq:Outer minimization}). All models predicted correctly while perturbing the original image using this attack (as in \cite{ghorbani2019interpretation})}
  \label{fig:ablation main}
  \vspace{-3mm}
\end{figure*}

\vspace{-5pt}
\section{Experiments and Results}
\label{sec:Experiments}
We conducted a comprehensive suite of experiments and ablation studies, which we report in this section and in Sec \ref{sec:Ablation}. We report results with our method on 4 benchmark datasets i.e. Flower \cite{nilsback2006visual}, Fashion-MNIST \cite{xiao2017fashion}, MNIST \cite{lecun2010mnist} and GTSRB \cite{stallkamp2012man}. The Flower dataset \cite{nilsback2006visual} contains 17 categories with each category consisting of 40 to 258 high-definition $128\times128$ RGB flower images. MNIST \cite{lecun2010mnist} and Fashion-MNIST \cite{xiao2017fashion} consist of $28\times28$ grayscale images from 10 categories of handwritten digits and fashion products respectively. GTSRB \cite{stallkamp2012man} is a physical traffic sign dataset with 43 classes and around 50,000 images in total. We compare the performance of our method against existing methods: RAR \cite{chen2019robust} for attributional robustness, Singh et al \cite{singh2019benefits}, and
Madry et al \cite{madry2017towards} which uses only standard adversarial training. Note that \cite{singh2019benefits}'s code is not publicly available, and we hence compared their results only on settings reported in their paper.

\noindent \textbf{Architecture Details:} We used a network consisting of two convolutional layers with 32 and 64 filters respectively, each followed by $2 \times 2$ max-pooling, and a fully connected layer with 1024 neurons, for experiments with both MNIST and Fashion-MNIST datasets. We used the Resnet model in \cite{zagoruyko2016wide} to perform experiments with Flower and GTSRB datasets and performed per image standardization before feeding images to the network consisting of 5 residual units with (16, 16, 32, 64) filters each. 
We also compared our results with a recently proposed method \cite{singh2019benefits} using WRN 28-10 \cite{zagoruyko2016wide} architecture as used in their paper. More architecture details for each dataset are provided in the Appendix; on any given dataset, the architectures were the same across all methods used for fair comparison.

\noindent \textbf{Performance Metrics:}  Following \cite{chen2019robust}\cite{singh2019benefits}, we used top-$k$ intersection, Kendall's correlation and Spearman correlation metrics to evaluate model's robustness against the IFIA attributional attack \cite{ghorbani2019interpretation} (Sec \ref{sec_bgd_prelims}). Top-$k$ intersection measures intersection size of the $k$ most important input features before and after the attributional attack. Kendall's and Spearman correlation compute rank correlation to compare the similarity between feature importance, before and after attack. We also report natural accuracy as well as adversarial accuracy, the latter being a metric to evaluate adversarial robustness of our model against adversarial attack, such as PGD (as described in eq.\ref{eq:PGD}). Here, adversarial accuracy refers to the accuracy over the adversarial examples generated from perturbations on the original test set using PGD (Eqn \ref{eq:PGD}).

We used a regularizer coefficient $\lambda=1.0$ and $m=50$ as the number of steps used for computing IG (Eqn \ref{eqn:ig}) across all experiments. Note that our adversarial and attributional attack configurations were kept fixed across ours and baseline methods. Please refer the Appendix for more details on training hyperparameters and attack configurations for specific datasets.

\begin{table*}[h]
\footnotesize
    \begin{minipage}{.5\linewidth}
        \begin{center}
		\begin{tabular}{|p{1.35cm}|p{1.10cm}|p{1.20cm}|p{0.9cm}|p{0.95cm}|}
		\hline \hline
			Dataset & Nat. acc.& Adv. acc.&
			Top-K & Kendall \\
			\hline \hline
            Flower & 83.09\% & 49.26\% & 67.95\% & 0.8012 \\
            \hline
            F-MNIST & 85.43\% & 71.25\% & 78.77\% &  0.6974\\
            \hline
            MNIST & 98.35\% &  88.66\% &  76.62\% &  0.3203\\
            \hline \hline
		\end{tabular}
		\label{tab1}
	\end{center}
	\vspace{-10pt}
		\caption{Results of using only $\mathcal{L}_{CACR}$}
		\label{tab:Ablation CCR}
    \end{minipage}%
    \begin{minipage}{.5\linewidth}
        \begin{center}
		\begin{tabular}{|p{1.35cm}|p{1.10cm}|p{1.20cm}|p{0.9cm}|p{0.95cm}|}
		\hline \hline
			Dataset & Nat. acc.& Adv. acc.&
			Top-K & Kendall \\
			\hline \hline
            Flower & 82.35\% & 50.00\% & 68.41\% & 0.8065 \\
            \hline
            F-MNIST & 85.44\% & 71.32\% & 79.11\% &  0.7000\\
            \hline
            MNIST & 98.37\% &  88.78\% &  77.88\% &  0.3217\\
            \hline \hline
		\end{tabular}
		\label{tab1}
	\end{center}
	    \vspace{-10pt}
		\caption{Results of using only $\mathcal{L}_{WACR}$}
		\label{tab:Ablation BAWR}
    \end{minipage}
    \vspace{-3mm}
\end{table*}

\begin{table*}[!htb]
\footnotesize
    \begin{minipage}{.5\linewidth}
        \begin{center}
		\begin{tabular}{|p{1.35cm}|p{1.10cm}|p{1.20cm}|p{0.9cm}|p{0.95cm}|}
		\hline \hline
			Dataset & Nat. acc.& Adv. acc.&
			Top-K & Kendall \\
			\hline \hline
            Flower & 82.47\%  & 50.97\%  & 69.04  & 0.8101 \\
            \hline
            F-MNIST & 85.45\% & 71.56\% & 80.19\% &  0.7138\\
            \hline
            MNIST&  98.39\% & 89.61\% & 79.18\%  & 0.3337  \\
            \hline \hline
		\end{tabular}
	\end{center}
        \vspace{-8pt}
		\caption{Proposed method with regularizer coeff for $\mathcal{L}_{CACR}$=1   and $\mathcal{L}_{WACR}$=0.7}
		\label{tab:Effect CCR1 BAWR0.5}
    \end{minipage}%
    \begin{minipage}{.5\linewidth}
        \begin{center}
		\begin{tabular}{|p{1.35cm}|p{1.10cm}|p{1.20cm}|p{0.9cm}|p{0.95cm}|}
		\hline \hline
			Dataset & Nat. acc.& Adv. acc.&
			Top-K & Kendall \\
			\hline \hline
            Flower & 82.44\%  & 50.81\%  & 68.63\%  & 0.8087  \\
            \hline
            F-MNIST & 85.44\% & 71.43\% & 79.86\% &  0.7091\\
            \hline
            MNIST& 98.43\% & 89.59\%    & 78.83\%  & 0.3283 \\
            \hline \hline
		\end{tabular}
	\end{center}
	    \vspace{-8pt}
		\caption{Proposed method with regularizer coeff for $\mathcal{L}_{CACR}$=0.7 and $\mathcal{L}_{WACR}$=1}
		\label{tab:Effect CCR0.5 BAWR1}
    \end{minipage}
\end{table*}

\noindent \textbf{Results:}
Tables \ref{tab:Flower main}, \ref{tab:Fashion-MNIST main}, \ref{tab:MNIST main} and \ref{tab:GTSRB main} report comparisons of natural/normal accuracy, adversarial accuracy, median value of top-$k$ intersection measure (shown as Top-K) and median value of Kendall's correlation (shown as Kendall), as used in \cite{chen2019robust}, on test sets of Flower, Fashion-MNIST, MNIST and GTSRB datasets respectively. (Note that \cite{singh2019benefits} did not report results on these architectures, and we report comparisons with them separately in later tables.) 
Our method shows significant improvement in performance on the Top-K and Kendall metrics - the metrics for attributional robustness in particular - across these datasets. 
Natural and adversarial accuracies are expected to be the highest for natural training and adversarial training method, Madry et al \cite{madry2017towards} respectively, and this is reflected in the results. 
A visual result is presented in Fig \ref{fig:Introduction Image}. More such qualitative results are presented in the Appendix. We show the variations in top-K intersection value and Kendall's correlation over all test samples for all the aforementioned 4 datasets using our method and RAR \cite{chen2019robust} in Fig \ref{fig:All Box Results}. Our variance is fairly similar to the variance in RAR. 

Tables \ref{tab:Flower WRN} and \ref{tab:GTSRB WRN} report the performance comparison (same metrics) on the Flower and GTSRB datasets using the WRN 28-10 architecture and hyperparameter settings used in \cite{singh2019benefits}. Note that RAR doesn't report results with this architecture, and hence is not included. We outperform Singh et al \cite{singh2019benefits} by significant amounts on the Top-K and Kendall metrics, especially on the Flower dataset. A comparison with Tables \ref{tab:Flower main} and \ref{tab:GTSRB main} makes it evident that the use of the WRN 28-10 architecture leads to significant improvement in attributional robustness.



Our results vindicate the methodology proposed in this work for the state-of-the-art results obtained for attributional robustness. Although we have additional loss terms, our empirical studies showed an increase of atmost 20-25\% in training time over RAR. Note that at test time, which is perhaps more important in deployment of such models, there is no additional time overhead for our method.

\section{Ablation Studies and Analysis}
\noindent \textbf{Quality of Saliency maps:}
It is important that attributional robustness methods do not distort the explanations significantly. One way to measure the quality of the generated explanations is through the deviation of attrib maps before and after applying our method. To judge the quality of our saliency maps, we compared the attributions generated by our method with the attributions of the original image from a naturally trained model and report the Spearman correlation in Table \ref{tab:Attribution comparison} for all datasets. The results clearly show that our saliency maps change lesser from original ones than other methods. We also conducted a human Turing test to check the goodness of the saliency maps by asking 10 human users to pick a single winning saliency map that was most true to the object in a given image among (Madry, RAR, Ours). The winning rates (in same order) were: 
MNIST: [30\%,30\%,40\%]; FMNIST: [20\%,40\%,40\%]; GTSRB: [20\%,30\%,50\%]; Flower: [0\%,30\%,70\%], showing that our saliency maps were truer to the image than other methods, especially on more complex datasets.\\

\begin{table}[t]
\centering
\vspace{-4mm}
\footnotesize
	\begin{tabular}{|p{1.35cm}||p{1.35cm}|p{1.35cm}|p{1.35cm}|}
		\hline \hline
			Dataset & Madry & RAR & OURS \\
		\hline
		Flower & 0.7234 & 0.8015 & \textbf{0.9004} \\
        \hline
        F-MNIST & 0.7897 & 0.8634 & \textbf{0.9289} \\
        \hline
		MNIST & 0.9826 & 0.9928 & \textbf{0.9957} \\
        \hline
        GTSRB & 0.8154 & 0.8714 & \textbf{0.9368} \\
        \hline \hline
	\end{tabular}
    \vspace{-8pt}
	\caption{Spearman correlation between attributions from diff methods w.r.t. attribs from a naturally trained model}
	\label{tab:Attribution comparison}
	\vspace{-4mm}
\end{table}

\vspace{-8pt}
\noindent \textbf{Effect of Regularizers:}
We analyzed the effect of each regularizer term which we introduced in the outer minimization formulation in Eqn \ref{Eq:Outer minimization}. For all such studies, the inner maximization setup was kept fixed. We compared attributional and adversarial accuracies, median values of Top-K intersection and Kendall's correlation achieved with and without $\mathcal{L}_{CACR}$ and $\mathcal{L}_{WACR}$ in the outer minimization. The results reported in Tables \ref{tab:Ablation CCR} and \ref{tab:Ablation BAWR} suggest that the performance deteriorated substantially by removing either $\mathcal{L}_{CACR}$ or $\mathcal{L}_{WACR}$, when compared to our original results in Tables \ref{tab:Flower main},\ref{tab:Fashion-MNIST main} and \ref{tab:MNIST main} for Flower, Fashion-MNIST and MNIST datasets. 

Fig \ref{fig:ablation main} shows the same effect visually with a sample test image from the Flower dataset. $\mathcal{L}_{CACR}$ not only captures the effect of positive class, but also diminishes the effect of most confusing negative class. Absence of $\mathcal{L}_{CACR}$ may hence consider attributions towards pixels which don't belong to the positive class. We can see that removing $\mathcal{L}_{CACR}$ increased the focus on a leaf which is not the true class (flower) in Fig \ref{fig:ablation main}(b) as compared to Fig \ref{fig:ablation main}(a).
$\mathcal{L}_{WACR}$ penalized a large attribution change on true class pixels (i.e. pixels with high base attribution).
This can be viewed from images in Fig \ref{fig:ablation main}(b) where keeping $\mathcal{L}_{CACR}$ forces minimal attribution change to true class pixels, compared to pixels outside the true class.
Fig \ref{fig:ablation main}(c) shows the result of using both regularizers which shows the best performance. More such qualitative results are also provided in the Appendix.\\



\vspace{-8pt}
\noindent \textbf{Effect of Reqularizer Coefficients:}
To investigate the relative effects of each proposed regularizer term, we performed experiments with other choices of regularizer coefficients. Tables \ref{tab:Effect CCR1 BAWR0.5} and \ref{tab:Effect CCR0.5 BAWR1} show the results. Our results suggest that the performance on attributional robustness drops for both cases across all datasets, when $\mathcal{L}_{CACR}$ and $\mathcal{L}_{WACR}$ are weighted lesser (original experiments had both weights to be 1). The drop is slightly more when $\mathcal{L}_{CACR}$ is weighted lesser, although this is marginal. 

Additional ablation studies, including the effect of the $KL(P_{ME} || P_{NCA})$ in Eqn \ref{eq_CACR_computation} and the use of Base Attribution in $\mathcal{L}_{WACR}$, are included in the Appendix due to space constraints.

\vspace{-5mm}
\section{Conclusions}
In this paper, we propose two novel regularization techniques to improve robustness of deep model explanations through axiomatic attributions of neural networks. Our experimental findings show significant improvement in attributional robustness measures and put our method ahead of existing methods for this task. Our claim is supported by quantitative and qualitative results on several benchmark datasets, followed by earlier work. Our future work includes incorporating spatial smoothing on the attribution map generated by true class, which can provide sparse and localized heatmaps. We hope our findings will inspire discovery of new attributional attacks and defenses which offers a significant pathway for new developments in trustworthy machine learning.

\noindent\textbf{Acknowledgement :}This work has been partly supported by the funding received from MHRD, Govt of India, and Honeywell through the UAY program (UAY/IITH005). We also acknowledge IIT-Hyderabad and JICA for provision of GPU servers for the work. We thank the anonymous reviewers for their valuable feedback that improved the presentation of this work.

\bibliography{main}

\newpage
\clearpage
\appendix
\section*{APPENDIX: Enhanced Regularizers for Attributional Robustness}
In this appendix, we provide details that could not be included in the main paper owing to space constraints, including: (i) description of our method as an algorithm (for convenience of understanding); (ii) an analysis of how our method is connected to the inner maximization of adversarial robustness in \cite{madry2017towards}; (iii) detailed descriptions of hyperparameter settings and attack configurations for every dataset; (iv) additional results, including results with Spearman correlation, effect of $KL(P_{ME} || P_{NCA})$ term, and use of Base Attribution in $\mathcal{L}_{WACR}$; (v) More visualizations of effect of different regularization terms; as well as (vi) More qualitative results on all datasets considered, and comparisons with RAR \cite{chen2019robust}.

\section{Proposed Algorithm}
The proposed method is presented as Algorithm \ref{alg:Algorithm} for convenience of understanding.

\section{Analysis of Proposed Method in terms of Robust Prediction Objective in \cite{madry2017towards}}
It is possible to show that our method is connected to the widely used robust prediction objective in \cite{madry2017towards}. To this end, we begin by observing that since the proposed method is a training method (and not an attribution method), all axioms of attribution will hold for the Integrated Gradient (IG) approach used in our method by default, as for any other trained model. 
Thus, using the completeness property of IG, the sum of pixel attributions equals the predicted class probability (for a class under consideration). Since the positive class probability will be greater than the negative class probability for a model prediction, this implies $\mathcal{L}_{CACR} \geq 0$ (from Eqns \ref{eq_P_TCA_computation},\ref{eq_P_NCA_computation},\ref{eq_CACR_calculation}). Similarly, from Eqn \ref{eq_WACR_computation}, we have $\mathcal{L}_{WACR} \geq 0$, as $S(\cdot)$ is the $L_1$-norm. Also we have $l_{CE}(\textbf{x},\textbf{y};\theta)$ (cross-entropy loss) is $\mathcal{L}_{TCI}(\textbf{x})$ itself, as defined in the work (Eqn \ref{Eq:Inner maximization}). Hence, considering $\lambda=1$, our loss function can be written as:\\ 
\begin{multline}
    \big(\mathcal{L}_{CACR}+\mathcal{L}_{WACR} \big)+\max_{\textbf{x}^{\prime} \in N(\textbf{x},\epsilon)}  (l_{CE}(\textbf{x}^{\prime},\textbf{y};\theta)+S(IG^{l_{CE}}_{\textbf{x}}(\textbf{x},\textbf{x}^{\prime})))\\
    \geq \max_{\textbf{x}^{\prime} \in N(\textbf{x},\epsilon)}  (l_{CE}(\textbf{x}^{\prime},\textbf{y};\theta)+S(IG^{l_{CE}}_{\textbf{x}}(\textbf{x},\textbf{x}^{\prime})))\\
    \geq l_{CE}(\textbf{x}^{\prime},\textbf{y};\theta)+\max_{\textbf{x}^{\prime} \in N(\textbf{x},\epsilon)} (S(IG^{l_{CE}}_{\textbf{x}}(\textbf{x},\textbf{x}^{\prime})))\\
    = l_{CE}(\textbf{x}^{\prime},\textbf{y};\theta)+\max_{\textbf{x}^{\prime} \in N(\textbf{x},\epsilon)} (|l_{CE}(\textbf{x}^{\prime}) - l_{CE}(\textbf{x})|)\\
    \geq \max_{\textbf{x}^{\prime} \in N(\textbf{x},\epsilon)} l_{CE}(\textbf{x}^{\prime})
\end{multline}\\
\noindent which is the inner maximization of \cite{madry2017towards}. Thus, atrributional robustness can be viewed as an extension of the robust prediction objective typically used for adversarial robustness.

\begin{algorithm}[!h]
\small
\SetAlgoLined
\KwIn{Training data $\textbf{X} = {(\textbf{x}_{1},\textbf{y}_{1}), (\textbf{x}_{2},\textbf{y}_{2}), ..., (\textbf{x}_{N},\textbf{y}_{N})}$ , a baseline image $\textbf{x}_{0}$ with all zero pixels, regularizer coefficient $\lambda$, step size of iterative perturbation $\alpha$, training batch size of $b$, number of epochs $E$, Classifier network $f$ parameterized by $\theta$.}
\KwOut{Parameters $\theta$ of classifier $f$.}
 Initialise parameters $\theta$.
 
 Compute $P_{ME} = \frac{1}{Number of Pixels}$.
 
 \For{epoch $\in {1,2,..,E}$}{
  Sample a batch $(\textbf{x},\textbf{y}) = { (\textbf{x}_{1},\textbf{y}_{1}), ..., (\textbf{x}_{b},\textbf{y}_{b}) }$
  
  Compute: $\mathcal{A}(\textbf{x}) = IG^{f_{TCI}}_{\textbf{x}}(\textbf{x}_{0}, \textbf{x})$, where $f_{TCI}$ represent classifier's true class output.(TCI = True Class Index)
  
  $\textbf{x}^{\prime} = \textbf{x} + Uniform[-\epsilon, + \epsilon]$
  
  \For{$i = 1,2,...,a$}{
   Compute Cross-Entropy loss, defined as $l_{CE}(\textbf{x}^{\prime},\textbf{y};\theta)$
   
   Compute $\nabla\tilde{\mathcal{A}} = IG^{\mathcal{L}_{TCI}}_{\textbf{x}}(\textbf{x}, \textbf{x}^{\prime})$, where $\mathcal{L}_{TCI}$ is the cross-entropy loss for the true class.
  
   $\textbf{x}^{\prime} = \textbf{x}^{\prime} + \alpha * sign(\nabla_{\textbf{x}} (l_{CE}(\textbf{x}^{\prime},\textbf{y};\theta) + S(\nabla\tilde{\mathcal{A}})))$ (Eqn. \ref{Eq:Inner maximization}), where we choose $S$ as L1-norm size function.
   
   $\textbf{x}^{\prime} = Proj_{l_{\infty}}(\textbf{x}^{\prime})$
   
   }
   
   Compute:  $P_{NCA} = \text{softmax}(IG^{f_{NCI}}_{\textbf{x}}(\textbf{x}_{0}, \textbf{x}^{\prime}))$, where $f_{NCI}$ represents classifier's most confusing negative class output.(NCI = $\operatorname*{argmax}_{j\neq TCI} (logit_{j})$)
   
   Compute: $P_{TCA} = \text{softmax}\big(IG^{f_{TCI}}_{\textbf{x}}(\textbf{x}_{0}, \textbf{x}^{\prime})\big)$
   
   Compute: $\mathcal{L}_{CACR} = KL(P_{ME}|| P_{NCA}) - KL(P_{ME}|| P_{TCA})$ (Eqn. \ref{eq_CACR_computation})
   
   Compute: $\nabla\mathcal{A}(\textbf{x}) = IG^{f_{TCI}}_{\textbf{x}}(\textbf{x}, \textbf{x}^{\prime}) = IG^{f_{TCI}}_{\textbf{x}}(\textbf{x}_{0}, \textbf{x}^{\prime}) - IG^{f_{TCI}}_{\textbf{x}}(\textbf{x}_{0}, \textbf{x})$
   
   Compute: $\mathcal{L}_{WACR} = S(\nabla\mathcal{A})\bigg|_{p \in P_1} + S(\mathcal{A} \odot \nabla\mathcal{A})\bigg|_{p \in P_2}$ (Eqn. \ref{eq_WACR_computation}), where P1 is set of pixels for sign$(\mathcal{A})$ $\neq$ sign$(\nabla\mathcal{A})$, P2 is set of pixels for sign$(\mathcal{A})$ = sign$(\nabla\mathcal{A})$ and $S$ is L1-norm size function.
    
    Compute: loss =  $l_{CE}(\textbf{x}^{\prime},\textbf{y};\theta) + \lambda \big(\mathcal{L}_{CACR} + \mathcal{L}_{WACR} \big)$ (Eqn. \ref{Eq:Outer minimization}) 
    
    Update $\theta$ which minimizes above loss.
   
 }
 \caption{Our Proposed Algorithm for Attributional Robustness}
 \label{alg:Algorithm}
\end{algorithm}

\section{Hyperparameters and Attack Configurations}
\label{sec:Hyperparameter settings}

Herein, we present details of training hyperparameters as well as attack configuration for all our experiments. We set regularization coefficient as $\lambda=1.0$ in these experiments, and $m=50$, the number of steps in computing IG (from zero image to given image) across all experiments for all datasets. (Note that $m$ is different in the attack step due to computational overhead in inner maximization. The specific values of $m$ in the attacks are provided under each dataset below.) For attributional attack, we use Iterative Feature Importance Attacks (IFIA) proposed by \cite{ghorbani2019interpretation} (specific settings for each dataset described below). We set the feature importance function as Integrated Gradients (IG) and dissimilarity function D as Kendall’s rank order correlation across all datasets. 
Also, we kept adversarial and attributional attack configurations fixed while comparing the result with other baseline methods, for fairness of comparison. 


\subsection{Flower Dataset:}
\textbf{Training Hyperparameters: }We use momentum optimizer with weight decay, momentum rate 0.9, weight decay rate 0.0002, batch size 16 and training steps 90,000. We use a learning rate schedule as follows: the first 1500 steps have a learning rate of $10^{-4}$; after 1500 steps and until 70,000 steps have a learning rate of $10^{-3}$; after 70,000 steps have a learning rate of $10^{-4}$. We use PGD attack as an adversary with a random start, number of steps of 7, step size of 2, $m=5$ as the number of steps for approximating IG computation in the attack step and adversarial budget $\epsilon$ of 8.

\noindent \textbf{Attack Configuration for Evaluation: }For evaluating adversarial robustness, we use a PGD attack with number of steps of 40, adversarial budget $\epsilon$ of 8 and step size of 2. For attributional attack, we use IFIA's top-$k$ attack with $k=1000$, adversarial budget $\epsilon=8$, step size $\alpha=1$ and number of iterations $P=100$.

\begin{figure*}[t]
\centering
  \includegraphics[width= 0.9\textwidth,height=5cm]{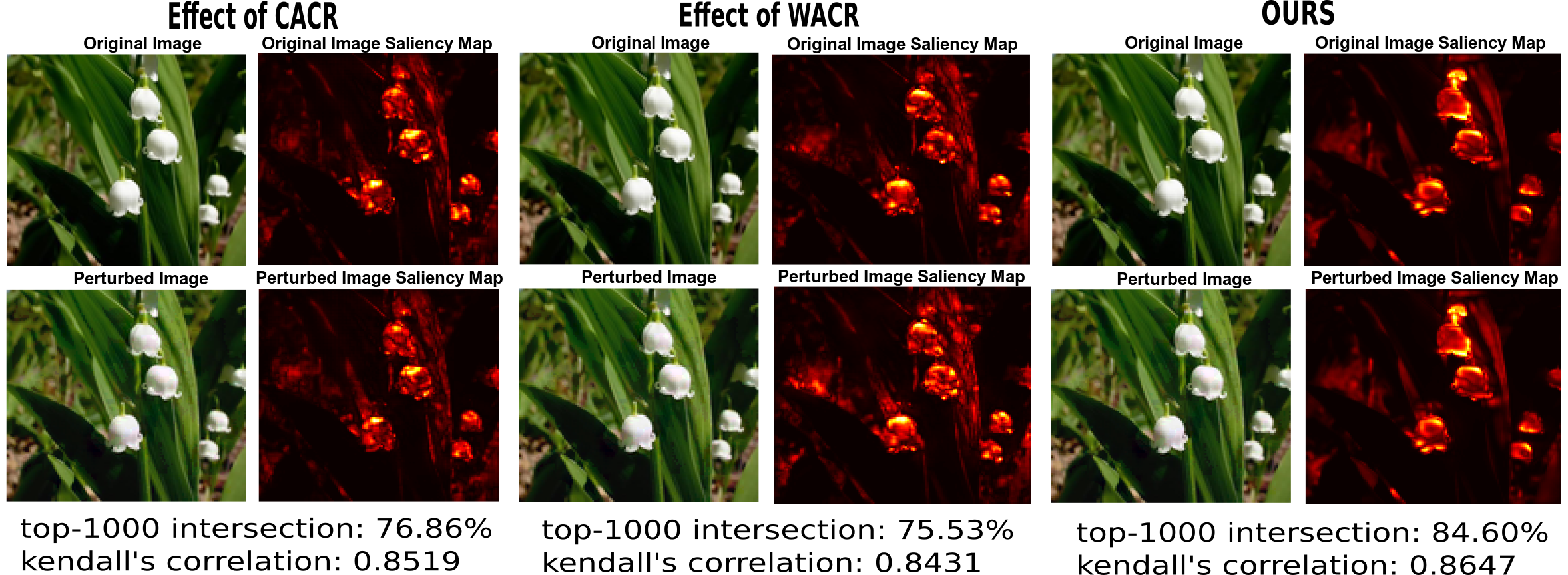}
  \caption{Visualizing effects of different regularization terms. Columns (a), (b) and (c) represent saliency maps of original and perturbed image considering only $\mathcal{L}_{CACR}$, only $\mathcal{L}_{WACR}$ and both regularizers respectively in outer minimization (Eqn \ref{Eq:Outer minimization}). We note that all models predicted correctly while perturbing the original image using this attack (as in \cite{ghorbani2019interpretation})}
  \label{fig:ablation flower visualization 1}
\end{figure*}

\begin{figure*}[t]
\centering
  \includegraphics[width= 0.9\textwidth,height=5cm]{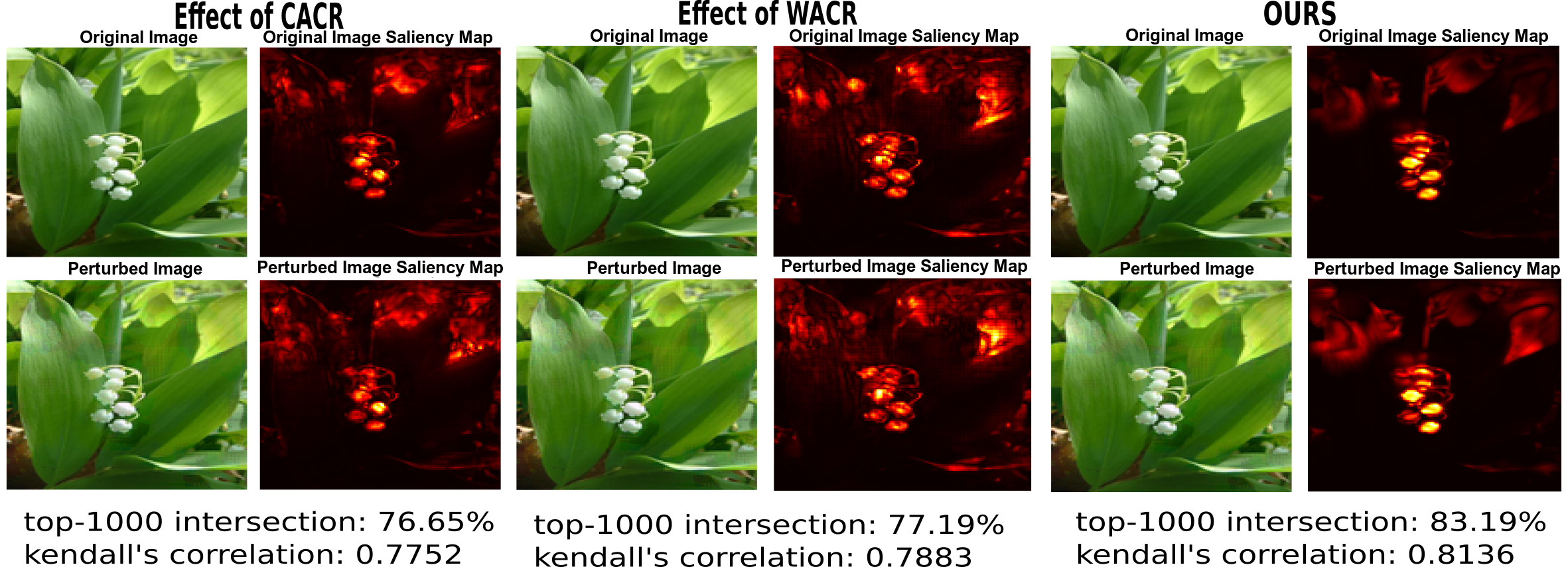}
  \caption{Visualizing effects of different regularization terms. Columns (a), (b) and (c) represent saliency maps of original and perturbed image considering only $\mathcal{L}_{CACR}$, only $\mathcal{L}_{WACR}$ and both regularizers respectively in outer minimization (Eqn \ref{Eq:Outer minimization}). We note that all models predicted correctly while perturbing the original image using this attack (as in \cite{ghorbani2019interpretation})}
  \label{fig:ablation flower visualization 2}
\end{figure*}

\subsection{Fashion-MNIST Dataset:}
\textbf{Training Hyperparameters: }We use learning rate as $10^{-4}$, batch size as 32, training steps as 100,000 and Adam optimizer. We use PGD attack as the adversary with a random start, number of steps of 20, step size of 0.01, $m=10$ as the number of steps for approximating IG computation in the attack step and adversarial budget $\epsilon=0.1$.

\noindent \textbf{Attack Configuration for Evaluation: }For evaluating adversarial robustness, we use PGD attack with random start, number of steps of 100, adversarial budget $\epsilon$ of 0.1 and step size of 0.01. For attributional attack, we use IFIA's top-k attack with $k=100$, adversarial budget $\epsilon=0.1$, step size $\alpha=0.01$ and number of iterations $P=100$.

\subsection{MNIST Dataset:}
\textbf{Training Hyperparameters: }We use learning rate as $10^{-4}$, batch size as 50, training steps as 90,000 and Adam optimizer. We use PGD attack as the adversary with a random start, number of steps of 40, step size of 0.01, $m=10$ as the number of steps for approximating IG computation in the attack step, and adversarial budget $\epsilon=0.3$.

\noindent \textbf{Evaluation Attacks Configuration: }For evaluating adversarial robustness, we use PGD attack with a random start, number of steps of 100, adversarial budget $\epsilon$ of 0.3 and step size of 0.01. For attributional attack, we use IFIA's top-$k$ attack with $k=200$, adversarial budget $\epsilon=0.3$, step size $\alpha=0.01$ and number of iterations $P=100$.

\subsection{GTSRB Dataset:}
\textbf{Training Hyperparameters: }We use momentum with weight decay rate 0.0002, momentum rate 0.9, batch size 32 and training steps 100,000. We use learning rate schedule as follows: the first 5000 steps have learning rate of $10^{-5}$; after 5000 steps and until 70,000 steps have learning rate of $10^{-4}$; after 70,000 steps have learning rate of $10^{-5}$. We use PGD attack as the adversary with a random start, number of steps of 7,step size of 2, $m=5$ as the number of steps for apprioximating IG computation in the attack step and adversarial budget $\epsilon=8$.

\noindent \textbf{Evaluation attacks Configuration:}For evaluating adversarial robustness, we use PGD attack with number of steps as 40, adversarial budget $\epsilon$ of 8 and step size of 2. For evaluating attributional robustness, we use IFIA's top-$k$ attack with $k=100$, adversarial budget $\epsilon=8$, step size $\alpha=1$ and number of iterations $P=50$.

\section{Additional Results}
\noindent \textbf{Spearman Correlation as Attributional Robustness Metric:} We compared our method with RAR \cite{chen2019robust} and a recently proposed attributional robustness method \cite{singh2019benefits} in the main paper. Singh et al \cite{singh2019benefits} also reported the median value of Spearman Correlation metric as an extra attributional robustness metric. We compare with them on this metric in Table \ref{tab:Ablation Spearman} below. Note that we outperform their method on this metric too. (We do not compare with RAR in this table, since RAR did not use WRN 28-10 architecture in their experiments, and RAR's numbers were significantly lower than those reported in the table below for this reason.)
\begin{table}[!htb]
\footnotesize
\centering
	\begin{tabular}{|p{1.95cm}|p{1.90cm}|p{1.90cm}|}
		\hline \hline
			Method & Spearman Corr(flower) & Spearman Corr(GTSRB) \\
    		\hline \hline
			Natural & 0.7413 & 0.9133 \\
            \hline
            Madry et al. & 0.9613 & 0.9770 \\
            \hline
            Singh et al. & 0.9627 & 0.9801 \\
            \hline
            Ours & \textbf{0.9959} & \textbf{0.9938} \\
            \hline \hline
	\end{tabular}
	\label{tab1}
	\caption{Comparison of median value of Spearman Correlation for different methods using WRN 28-10 architecture.}
	\label{tab:Ablation Spearman}
	\vspace{-3mm}
\end{table}

\begin{figure}[t]
\centering
  \includegraphics[width= 0.47\textwidth,height=2cm]{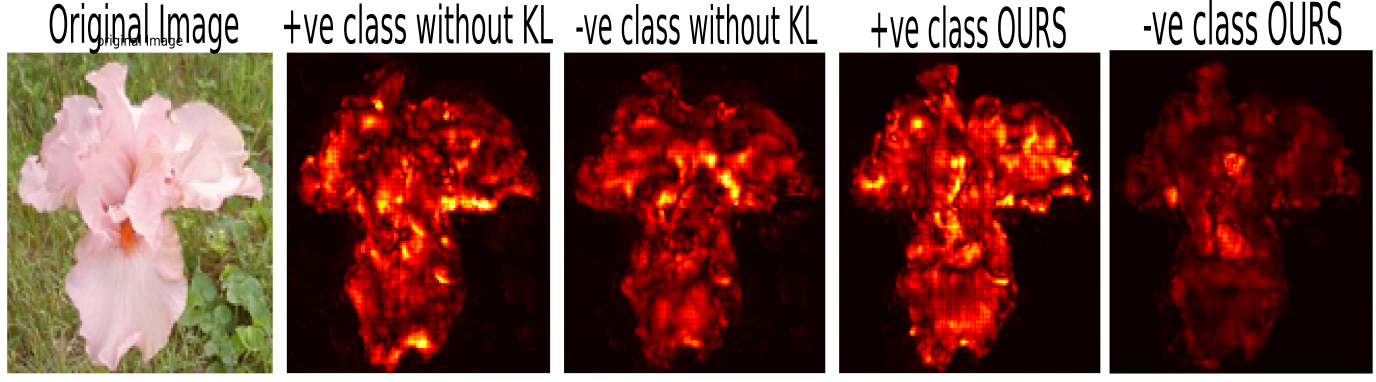}
  \vspace{-8pt}
  \caption{Visualizing effects of $KL(P_{ME} || P_{NCA})$}
  \label{fig:ablation KL}
\end{figure}

\noindent \textbf{Effect of $KL(P_{ME} || P_{NCA})$:} We studied the effect of the term $KL(P_{ME}||P_{NCA})$ by training a model without this term, keeping everything else the same i.e. calculating $\mathcal{L}_{CACR}$ without this term in Eqn \ref{eq_CACR_computation}. The images in Fig \ref{fig:ablation KL} show that using this term improves focus of attribution on true class pixels, reducing attribution of most confusing negative class.

\noindent \textbf{Use of Base Attribution in $\mathcal{L}_{WACR}$:} In order to understand the usefulness of weighting the change in attribution with the base attribution in $\mathcal{L}_{WACR}$ (Eqn \ref{eq_WACR_computation}), we replaced $S(\mathcal{A} \odot \nabla\mathcal{A})$ with $S(\nabla\mathcal{A})$ in Eqn \ref{Eq:Outer minimization} for the outer minimization step and trained our model keeping all other settings intact. Table \ref{tab:Ablation BAWR comparison} reports natural and adversarial accuracies, median values of Top-K intersection and Kendall's correlation achieved with the modified formulation, which shows substantial reduction on all results compared to our original results in tables \ref{tab:Flower main}, \ref{tab:Fashion-MNIST main} and \ref{tab:MNIST main}. This supports our claim for construction of $\mathcal{L}_{WACR}$.

\begin{table}[!htb]
\centering
\footnotesize
	\begin{tabular}{|p{1.35cm}|p{1.10cm}|p{1.20cm}|p{0.9cm}|p{0.95cm}|}
		\hline \hline
			Dataset & Nat. acc.& Adv. acc.&
			Top-K & Kendall \\
		\hline \hline
        Flower & 82.35\% & 50.74\% & 68.71\% & 0.8089 \\
        \hline
        F-MNIST & 85.43\% & 71.38\% & 79.89\% &  0.7023\\
        \hline
        MNIST& 98.41\% &  89.53\% &  79.00\% & 0.3315 \\
        \hline \hline
	\end{tabular}
	\label{tab1}
	\caption{Removing use of Base Attribution in $\mathcal{L}_{WACR}$ for outer minimization on Flower, FMNIST and MNIST}
	\label{tab:Ablation BAWR comparison}
	\vspace{-7mm}
\end{table}

\section{Visualization of Effect of Different Regularization Terms}
Continuing from the main paper, we present more qualitative results showing the effect of both proposed regularizers i.e. $\mathcal{L}_{CACR}$ and $\mathcal{L}_{WACR}$ individually, and the performance when using both of them. Figures \ref{fig:ablation flower visualization 1} and \ref{fig:ablation flower visualization 2} show these examples from the Flower dataset.

\section{More Qualitative Results}
\begin{figure*}[!htbp]
\centering
  \includegraphics[width= 0.75\textwidth,height=5.5cm]{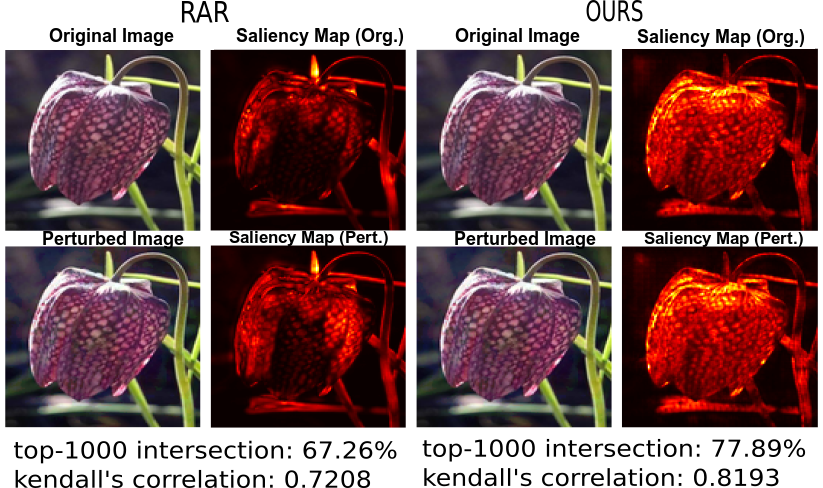}
  \caption{Original image and human-imperceptible attributionally attacked image (both correctly predicted) and their corresponding saliency maps under RAR \cite{chen2019robust} and our method for a test sample from Flower dataset.}
  \label{fig_flower_supplementary1}
\end{figure*}

\begin{figure*}[!htbp]
\centering
  \includegraphics[width= 0.75\textwidth,height=5.5cm]{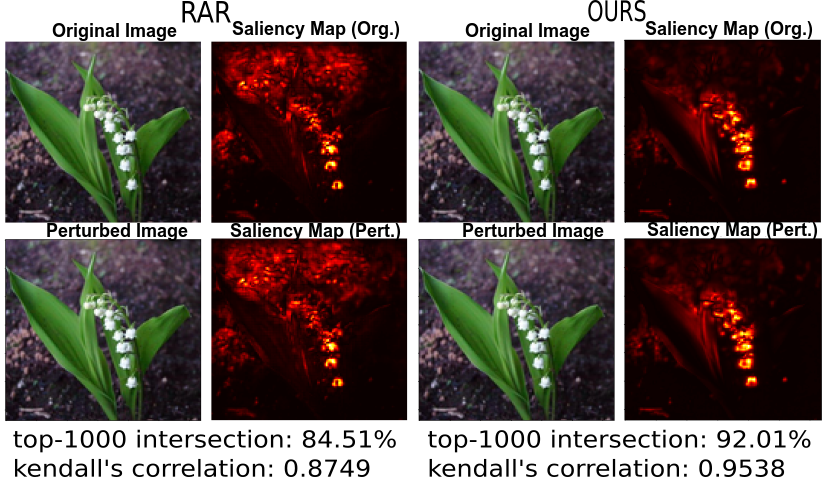}
  \caption{Original image and human-imperceptible attributionally attacked image (both correctly predicted) and their corresponding saliency maps under RAR \cite{chen2019robust} and our method for a test sample from Flower dataset.}
  \label{fig_flower_supplementary2}
\end{figure*}

\begin{figure*}[!htbp]
\centering
  \includegraphics[width= 0.75\textwidth,height=5.5cm]{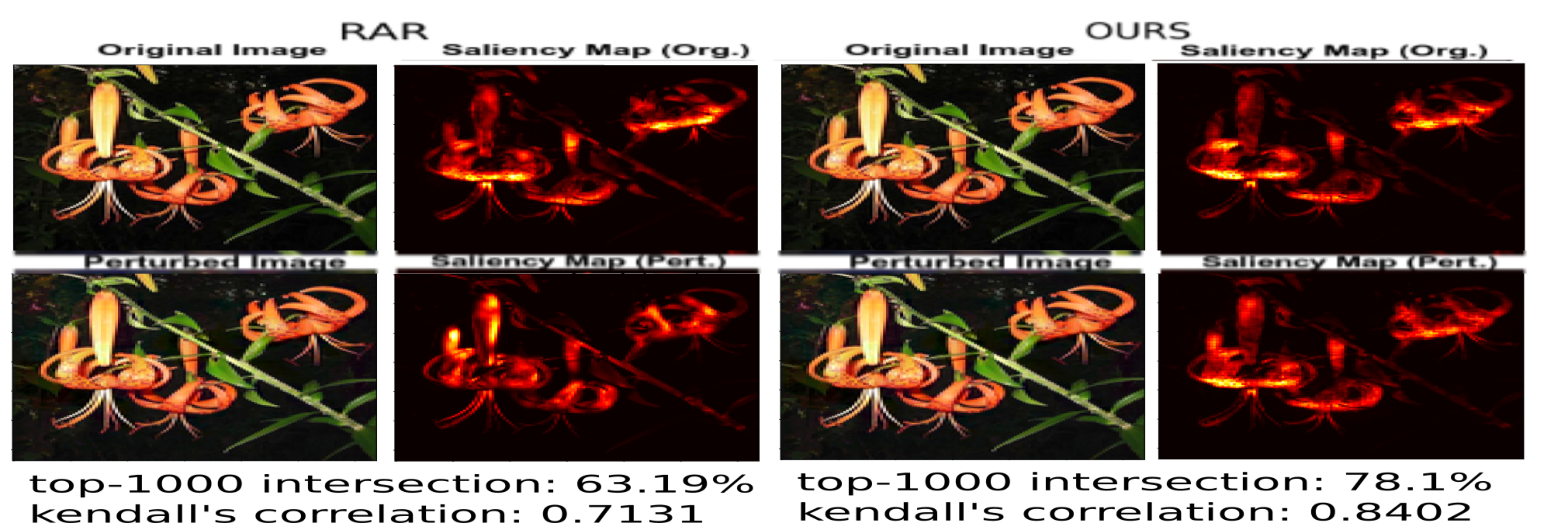}
  \caption{Original image and human-imperceptible attributionally attacked image (both correctly predicted) and their corresponding saliency maps under RAR \cite{chen2019robust} and our method for a test sample from Flower dataset.}
  \label{fig_flower_supplementary3}
\end{figure*}

\begin{figure*}[!htbp]
\centering
  \includegraphics[width= 0.75\textwidth,height=5.5cm]{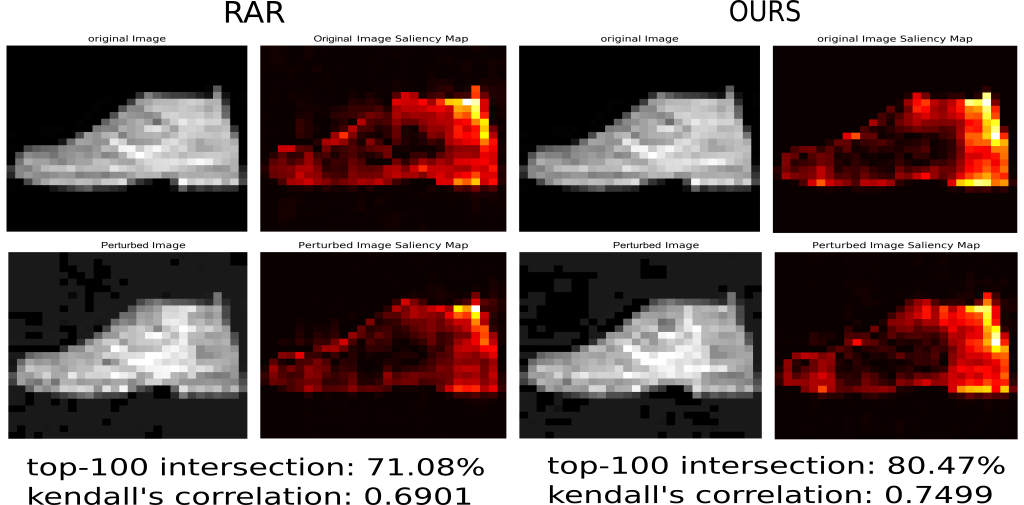}
  \caption{Original image and human-imperceptible attributionally attacked image (both correctly predicted) and their corresponding saliency maps under RAR \cite{chen2019robust} and our method for a test sample from Fashion-MNIST dataset.}
  \label{fig_fmnist_supplementary1}
\end{figure*}

\begin{figure*}[!htbp]
\centering
  \includegraphics[width= 0.75\textwidth,height=5.5cm]{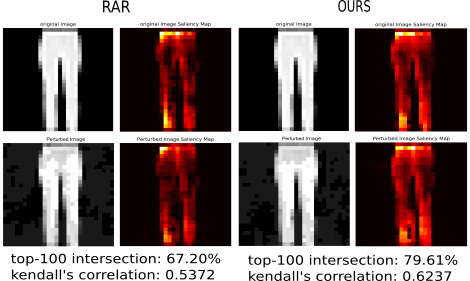}
  \caption{Original image and human-imperceptible attributionally attacked image (both correctly predicted) and their corresponding saliency maps under RAR \cite{chen2019robust} and our method for a test sample from Fashion-MNIST dataset.}
  \label{fig_fmnist_supplementary2}
\end{figure*}

\begin{figure*}[!htbp]
\centering
  \includegraphics[width=0.75\textwidth,height=5.5cm]{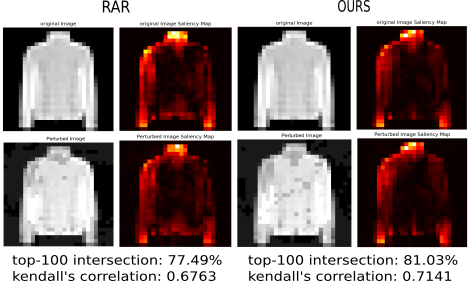}
  \caption{Original image and human-imperceptible attributionally attacked image (both correctly predicted) and their corresponding saliency maps under RAR \cite{chen2019robust} and our method for a test sample from Fashion-MNIST dataset.}
  \label{fig_fmnist_supplementary3}
\end{figure*}

\begin{figure*}[!htbp]
\centering
  \includegraphics[width= 0.75\textwidth,height=5.5cm]{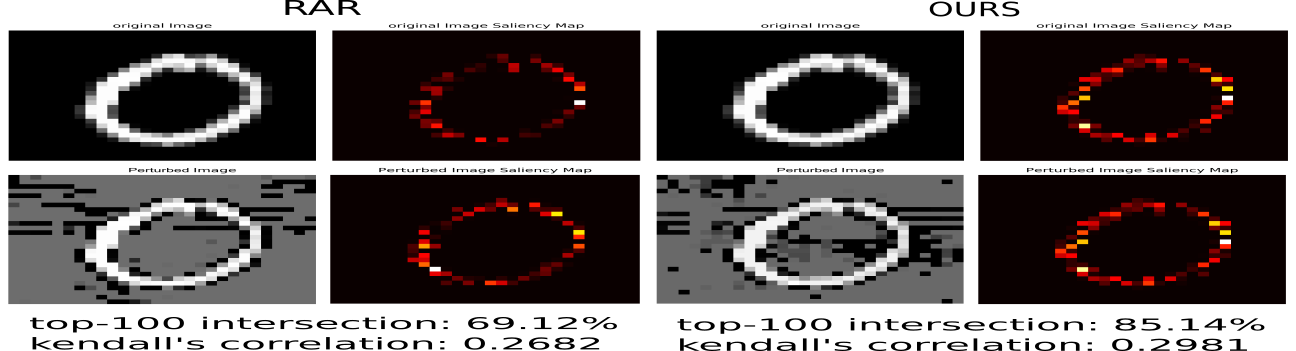}
  \caption{Original image and human-imperceptible attributionally attacked image (both correctly predicted) and their corresponding saliency maps under RAR \cite{chen2019robust} and our method for a test sample from MNIST dataset.}
  \label{fig_mnist_supplementary1}
\end{figure*}

\begin{figure*}[!htbp]
\centering
  \includegraphics[width=0.75\textwidth,height=5.5cm]{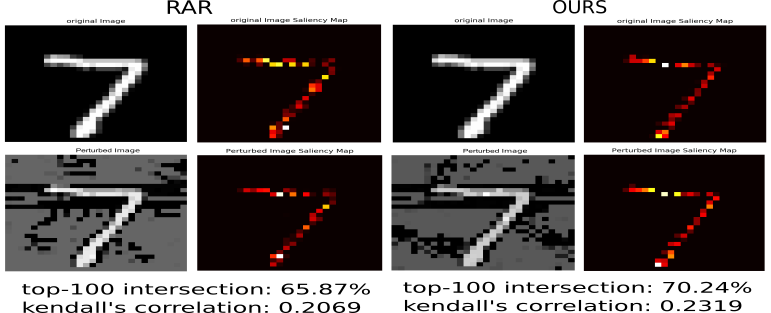}
  \caption{Original image and human-imperceptible attributionally attacked image (both correctly predicted) and their corresponding saliency maps under RAR \cite{chen2019robust} and our method for a test sample from MNIST dataset.}
  \label{fig_mnist_supplementary2}
\end{figure*}

\begin{figure*}[!htbp]
\centering
  \includegraphics[width= 0.75\textwidth,height=5.5cm]{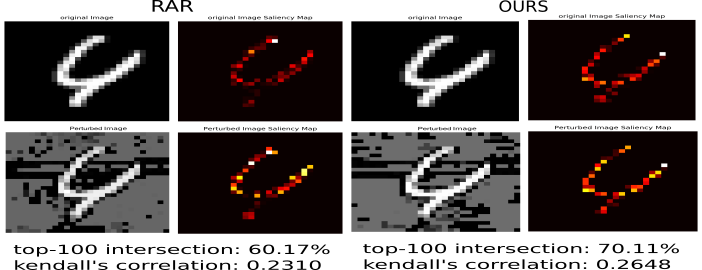}
  \caption{Original image and human-imperceptible attributionally attacked image (both correctly predicted) and their corresponding saliency maps under RAR \cite{chen2019robust} and our method for a test sample from MNIST dataset.}
  \label{fig_mnist_supplementary3}
\end{figure*}

\begin{figure*}[!htbp]
\centering
  \includegraphics[width=0.75\textwidth,height=5.5cm]{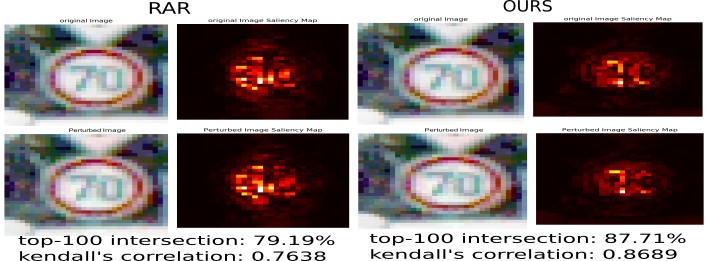}
  \caption{Original image and human-imperceptible attributionally attacked image (both correctly predicted) and their corresponding saliency maps under RAR \cite{chen2019robust} and our method for a test sample from GTSRB dataset.}
  \label{fig_gtsrb_supplementary1}
\end{figure*}

\begin{figure*}[!htbp]
\centering
  \includegraphics[width=0.75\textwidth,height=5.5cm]{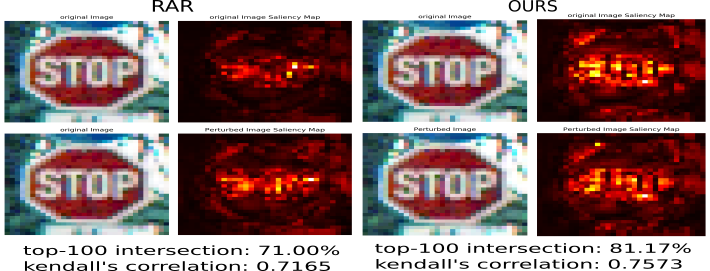}
  \caption{Original image and human-imperceptible attributionally attacked image (both correctly predicted) and their corresponding saliency maps under RAR \cite{chen2019robust} and our method for a test sample from GTSRB dataset.}
  \label{fig_gtsrb_supplementary2}
\end{figure*}

\begin{figure*}[!htbp]
\centering
  \includegraphics[width=0.75\textwidth,height=5.5cm]{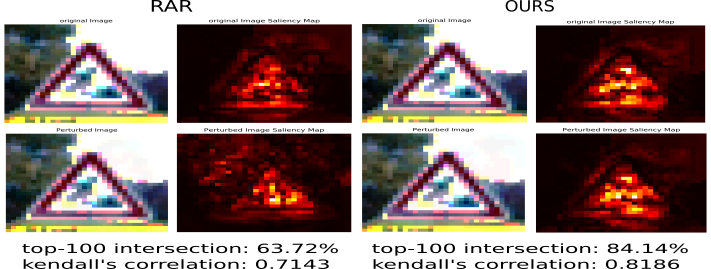}
  \caption{Original image and human-imperceptible attributionally attacked image (both correctly predicted) and their corresponding saliency maps under RAR \cite{chen2019robust} and our method for a test sample from GTSRB dataset.}
  \label{fig_gtsrb_supplementary3}
\end{figure*}

We now present more qualitative results from each dataset on the following pages, comparing our proposed robust attribution method with RAR. Figures \ref{fig_flower_supplementary1}-\ref{fig_gtsrb_supplementary3} (following pages) show these results. (Note that Singh et al \cite{singh2019benefits} do not have their code publicly available, and do not report these results - we hence are unable to compare with them on these results.) Note the consistent performance of our method across all these examples across datasets.

\end{document}